\documentclass[runningheads]{llncs}
\usepackage{graphicx}
\usepackage{multirow}
\usepackage{bbding}
\usepackage{float}
\newfloat{figtab}{htb}{fgtb}
\makeatletter
  \newcommand\figcaption{\def\@captype{figure}\caption}
  \newcommand\tabcaption{\def\@captype{table}\caption}
\makeatother

\begin{document}

\title{Human-machine Interactive Tissue Prototype Learning for Label-efficient Histopathology Image Segmentation}
\titlerunning{Human-machine Interactive Tissue Prototype Learning}
%
\author{Wentao Pan \inst{1} \and
 Jiangpeng Yan\inst{3}\inst{(}\Envelope\inst{)} \and
 Hanbo Chen \inst{2} \and Jiawei Yang \inst{4} \and Zhe Xu \inst{5} \and Xiu Li \inst{1}\inst{(}\Envelope\inst{)} \and Jianhua Yao \inst{2}\inst{(}\Envelope\inst{)}\thanks{W. Pan, J. Yan, and H. Chen contributed equally. J. Yan, X. Li and J. Yao are corresponding authors.}}
\authorrunning{W. Pan, et al.}

\institute{Tsinghua Shenzhen International Graduate School, Shenzhen, China \\ \email{li.xiu@sz.tsinghua.edu.cn} \and
 Tencent AI Lab, Shenzhen, China \\ \email{yaojianhua@tencent.com} \and
 Department of Automation, Tsinghua University, Beijing, China \email{yanjp13@tsinghua.org.cn} \\ \and
 University of California, Los Angeles, USA \\ \and
The Chinese University of Hong Kong, Hong Kong, China
 }

\maketitle              
\begin{abstract}
Deep learning have greatly advanced histopathology image segmentation but usually require abundant annotated data. However, due to the gigapixel scale of whole slide images and pathologists' heavy daily workload, obtaining pixel-level labels for supervised learning in clinical practice is often infeasible. Alternatively, weakly-supervised segmentation methods have been explored with less laborious image-level labels, but their performance is unsatisfactory due to the lack of dense supervision. Inspired by the recent success of self-supervised learning, we present a label-efficient tissue prototype dictionary building pipeline and propose to use the obtained prototypes to guide histopathology image segmentation. Particularly, taking advantage of self-supervised contrastive learning, an encoder is trained to project the unlabeled histopathology image patches into a discriminative embedding space where these patches are clustered to identify the tissue prototypes by efficient pathologists' visual examination. Then, the encoder is used to map the images into the embedding space and generate pixel-level pseudo tissue masks by querying the tissue prototype dictionary. Finally, the pseudo masks are used to train a segmentation network with dense supervision for better performance. Experiments on two public datasets demonstrate that our method can achieve comparable segmentation performance as the fully-supervised baselines with less annotation burden and outperform other weakly-supervised methods. Codes are available at https://github.com/WinterPan2017/proto2seg.

\keywords{WSI Segmentation \and Label-efficient Learning \and Clustering.}
\end{abstract}

\section{Introduction}
\begin{figure}[t]
\centering
\includegraphics[width=\columnwidth]{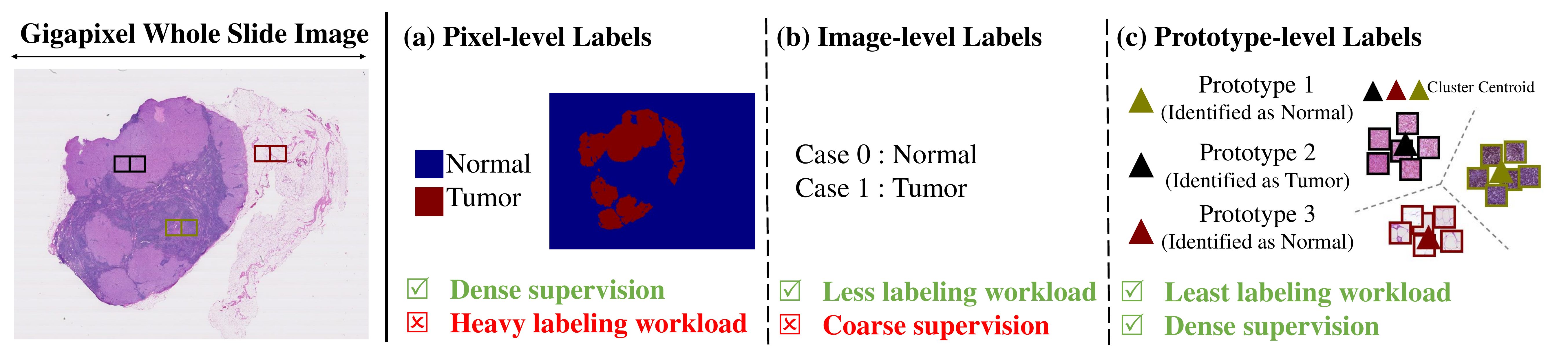}
\caption{Comparison of different WSI annotation types: (a) pixel-level, (b) image-level, and our (c) prototype-level labels. }
\label{fig_anno_cost}
\end{figure}

Visual examination of tissue sections under a microscope is a crucial step for disease assessment and prognosis. 
Automatic segmentation of WSIs is in high demand because it helps pathologists quantify tissue distribution. 
So far, supervised learning approaches have shown state-of-the-art performance on WSI segmentation with abundant annotated data \cite{chan2019histosegnet}. 
Unfortunately, it takes hours to annotate pixel-wise labels for a WSI with the gigapixel scale
as shown in Fig. \ref{fig_anno_cost} (a).
Thus, abundant pixel-level labels for supervised learning are often infeasible in busy daily clinical practice. Alternatively, weakly-supervised segmentation methods have been explored to supervise the model training process with less laborious image-level labels, as shown in Fig. \ref{fig_anno_cost} (b), and estimate segmentation results. However, these weakly-supervised methods suffer from unsatisfied pixel-level predictions due to the lack of dense supervision. Therefore, we wonder \textbf{if there exist other WSI analysis pipelines that can provide pixel-level supervision without a heavy labeling workload.} 

Recent years have witnessed substantial progress in the unsupervised learning of natural image analysis, especially that made by contrastive learning \cite{chen2020simple,grill2020bootstrap}. 
Inspired by previous works \cite{yang2021towards,yan2022deep} where researchers achieved high tissue classification accuracy by performing clustering algorithms on self-supervised learned histopathology representations, in this work, we make a further step to build a bridge between contrastive learning based WSI patch pre-training and pixel-level tissue segmentation with a human-machine interactive tissue prototype learning pipeline, namely Proto2Seg. Particularly, taking advantage of self-supervised contrastive learning, we crop unlabeled WSIs into local patches and use them to train an encoder that can project these patches into a discriminative embedding space. These patches are then divided into different clusters in the embedding space via unsupervised clustering. By examining dozens of representative tissue patches in each cluster as shown in Fig. \ref{fig_anno_cost} (c), pathologists can efficiently determine whether a cluster is a target tissue type or not. The centroids of pathologist-selected clusters are collected to build a tissue prototype dictionary, with which we can use the encoder to map the original WSIs into the embedding space and generate pseudo tissue masks by querying the nearest prototype to current local regions with flexible settings. We further adopt a refinement strategy to use the generated masks as the dense supervision for training a segmentation network from scratch. 
As such, pathologists only need to spend several minutes examining representative patches in every cluster for WSI segmentation model training rather than hour-counted dense annotation.

In summary, our contributions are in the following aspects: (1) We make one of the early attempts to bridge the contrastive learning-based WSI patch pre-training and dense segmentation by a low-labor-cost human-machine interactive labeling tissue prototype dictionary. (2) We propose an effective framework, namely Proto2Seg, to generate coarse tissue masks by querying the tissue prototype dictionary and designing a customized query process to further improve the coarse segmentation results. (3) By using the coarse tissue masks to supervise the training of histopathology segmentation networks, the quantitative and qualitative experimental results on two public datasets demonstrate that our approach achieves comparable segmentation performance to the fully-supervised upper bound and is superior to other weakly-supervised methods. 

\section{Related Works}
\noindent\textbf{Self-supervised WSI Analysis:} Inspired by the recent success of contrastive learning \cite{chen2020simple,grill2020bootstrap} in natural image analysis, there have been some works \cite{yang2021towards,yang2021self} where researchers fine-tuned a model with contrastive learning based pre-trained weights under image-level supervision for WSI patch classification task. 
That is, manual labels are still required in the fine-tuning process in these works, although the model is pre-trained in a self-supervised fashion.  
In a recent work \cite{yan2022deep}, researchers made an early attempt to distinguish different tissues with a recursive clustering algorithm, proving that integrating clustering with the contrastive learning-based histopathology representations can achieve high patch classification accuracy.
In contrast to the existed works, we focus on building a bridge between the clustering-based tissue classification and the pixel-level WSI segmentation to move the topic forward for label-efficient WSI analysis. 
 
\noindent\textbf{Weakly-supervised WSI Segmentation:} Since it is difficult to obtain pixel-level labels for WSI segmentation, weakly-supervised WSI segmentation has been an active research area for years. 
The weakly-supervised WSI segmentation methods can be roughly categorized into CAM-based and MIL-based solutions. 
The CAM-based methods are built on Class Activation Map variants \cite{zhou2016learning,selvaraju2017grad,chattopadhay2018grad} produced by well-trained classification models. Generally, these predictions are error-prone and need to be refined by complex post-processing strategies \cite{chan2019histosegnet}.  
MIL-based solutions \cite{ilse2018attention,xu2019camel,lu2021data} regard WSIs as a bag of local patches and can learn to predict the classification label of each local patch. The patch classification results are then merged as segmentation masks. Most MIL-based methods \cite{ilse2018attention,xu2019camel} are designed for binary segmentation scenario. Different from these works, our method can handle both binary and multiple tissue segmentation tasks with effective human-machine interactive steps and better results. 

\section{Methodology}
\label{sec:method}
\begin{figure}[t]
\centering
\includegraphics[width=1\columnwidth]{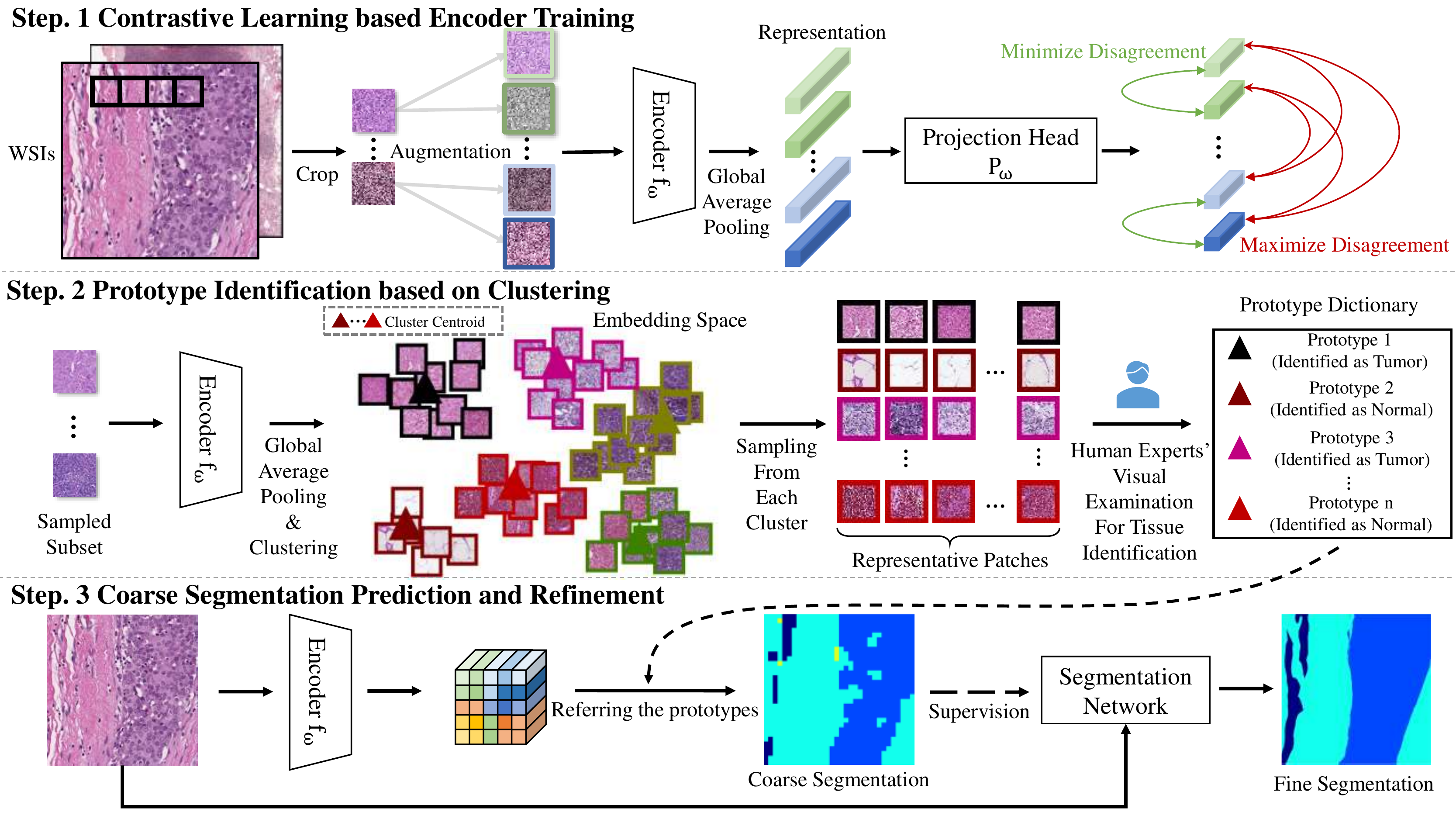}
\caption{Illustration of three main steps of our framework: 1) contrastive learning based encoder training (Machine), 2) prototype identification based on clustering (Machine + Human), 3) coarse segmentation prediction and refinement (Machine).}
\label{fig_framework}
\end{figure}

Our framework, as shown in Fig. \ref{fig_framework}, includes three steps: (1) contrastive learning-based encoder training, (2) prototype identification based on clustering, and (3) coarse segmentation prediction and refinement.

\subsection{Contrastive Learning based Encoder Training}

When pathologists read WSIs, they can recognize different tissues based on comparing local cells' visual appearance and surrounding micro-environments. Similarly, we need an encoder that can project local visual histopathology patterns into a discriminate space to identify different tissues. Inspired by recent studies on contrastive learning, we adopt SimCLR \cite{chen2020simple} to pre-train the encoder $f_\omega(\cdot)$ without labels in a self-supervised fashion. 

The detail of this step is shown in Fig. \ref{fig_framework} Step 1. Given a set of WSIs $\mathcal{I} = \{i_1, i_2, ..., i_m\}$, we crop them into non-overlap patches $\mathcal{P} = \{p_1, p_2, ..., p_n\} (n \gg m)$ with small resolution (empirically set as 128$\times$128 in this study). During the training stage, given N patches in a mini-batch, we get 2N patches by applying different augmentations on each patch. Two augmented patches from the same patch are regarded as positive pairs, others are treated as negative pairs. 
We employ ResNet18 \cite{he2016deep} backbone, where the final linear classification layer and global average pooling layer are removed, as the encoder $f_\omega(\cdot)$. 
Each patch $p_i \in \mathcal{R}^{128 \times 128 \times 3}$ is encoded by $f_\omega(\cdot)$ to get $h_i=f_\omega(p_i), h_i \in \mathcal{R}^{4 \times 4 \times 512} $, and then feed into global average pooling layer $avg(\cdot)$ to generate an embedding space with a dimension of $\mathcal{R}^{512}$. Same as \cite{chen2020simple}, a projection head $g(\cdot)$ is used to map $h_i$ to $z_i=g(avg(h_i)), z_i \in \mathcal{R}^{128}$ where the contrastive loss is applied. To achieve our goal, the contrastive loss is utilized to pull the positive pairs close and push the negative pairs away in the embedding space. Given a sample‘s embedding $s$ and $t^+, t^-$ as its positive and negative samples’ embeddings, the formula of contrastive loss is defined as: $L_{s,t^+,t^-} = - \log \frac{\exp (\frac{sim(s, t^+)}{\tau})}{\exp (\frac{sim(s, t^+)}{\tau})+\sum_{k^-} \exp (\frac{sim(s, t^-)}{\tau})}$, where $sim(\cdot)$ denotes the similarity function and $\tau$ denotes the temperature parameter. 
In practice, we utilize cosine similarity $cosine(x,y)=\frac{x \cdot y}{|x||y|}$ and set $\tau$ to 0.5. 
After training, $f_\omega(\cdot)$ is used to project histopathology patches into a discriminative embedding space of $\mathcal{R}^{512}$ for prototype identification.

It is reminded that except for SimCLR \cite{chen2020simple}, other self-supervised pre-training strategies \cite{he2020momentum,grill2020bootstrap} can be adopted to obtain the encoder. Here, we use SimCLR to demonstrate the effectiveness of our framework and leave how different self-supervised strategies can affect the results as a future work.

\subsection{Prototype Identification based on Clustering}
\label{sec:prototype_iden}
Having obtained a discriminative histopathology image encoding latent space, we then use clustering to mine potential tissue prototypes (i.e. cluster centroids), which can capture inter- and intra-class heterogeneity of different tissues and represent the entire encoding space in a dictionary. To map these prototypes with corresponding labels, only in this step do we need to integrate experts' examination. In practice, we can ask an expert to examine sampled representative patches to efficiently determine the labels of corresponding clusters without laboriously evaluating all the patches as shown in Fig. \ref{fig_framework} Step 2. 
In our experiments, we simulate the visual inspection process with ground-truth labels for easy reproducibility and evaluation of the segmentation performance. 

To begin with, we illustrate how to perform the clustering with the pre-trained $f_\omega(\cdot)$. To reduce computational overhead, we randomly sample a subset $\mathcal{P}_{sub}$ from the entire histopathology patch set $\mathcal{P}$. Then, $f_\omega(\cdot)$ and $avg(\cdot)$ are utilized to project patches $\mathcal{P}_{sub}$ into the discriminative embedding space to obtain the embedding set $\mathcal{E}_{sub}$. Next, we adopt unsupervised clustering on $\mathcal{E}_{sub}$ to find tissue prototypes without accessing any manual label. For computation efficiency, K-Means++ \cite{arthur2006k} is employed to generate $k$ clusters from $\mathcal{E}_{sub}$. Naturally, the first problem comes: \textbf{how to determine a proper cluster number $k$}?

Intuitively, the selection of $k$ plays a trade-off between reducing annotation workload and purifying clusters, i.e., a larger $k$ results in smaller tissue clusters with higher inter-class similarity, but costs more labor as the pathologist needs to inspect. In order to select a good cluster number without accessing any manual labels, 
we adopt the elbow method \cite{liu2020determine} to determine proper $k$ where model performance increment is no longer worth additional cost. 
We use an annotation-free distance-based measurement: inter-class embedding squared distance $D$ to measure the model performance. $D$ is given by:
$D = \sum_{j}^{k}{\sum_{*}{||centroid_{j}, p_{(j, *)}||_2}}$,
where $centroid_j$ denotes the centroid of cluster $j$ and $p_{(j, *)}$ denotes the embeddings of patches in cluster $j$. Generally, $D$ decreases when $k$ grows up. To search a proper $k$, we can plot the curve of relative reduction $R_v = D_{v-1}-D_{v}, v \geq 2$, where $v$ is potential values of $k$ (e.g., $v \in [2,3,...,50]$), and choose the elbow point where the reduction of $D$ slows down as the final cluster number.

It is noted that identifying proper cluster number is an open problem, and we also evaluate other clustering methods. Our experiments show that KMeans++ is better with current setting in our following ablation study. 

Based on the obtained clusters, then there comes the second problem: \textbf{how can human experts efficiently determine the tissue types for every cluster}? We propose to let human experts only evaluate $t$ sampled representative patches to give the prototype-level labels instead of inspecting all the patches. The $t$ representative patches are sampled by the central sampling strategy. That is, given a target cluster, we sample $t$ patches which are closest to the cluster centroid. More sampling strategies will be discussed in the experiments. To simulate pathologists' the visual examination process, here we count the proportion of each tissue according to the pixel-level labels. If there exists one tissue type with a proportion of pixels exceeding 80\%, we consider this cluster to represent this tissue and add the prototype (the cluster centroid embedding) into the prototype dictionary. Otherwise, we assume the cluster may be a mixture of different tissue types (i.e. the border area between organs), and drop it. 

\begin{figure*}[t]
\centering
\includegraphics[width=1\columnwidth]{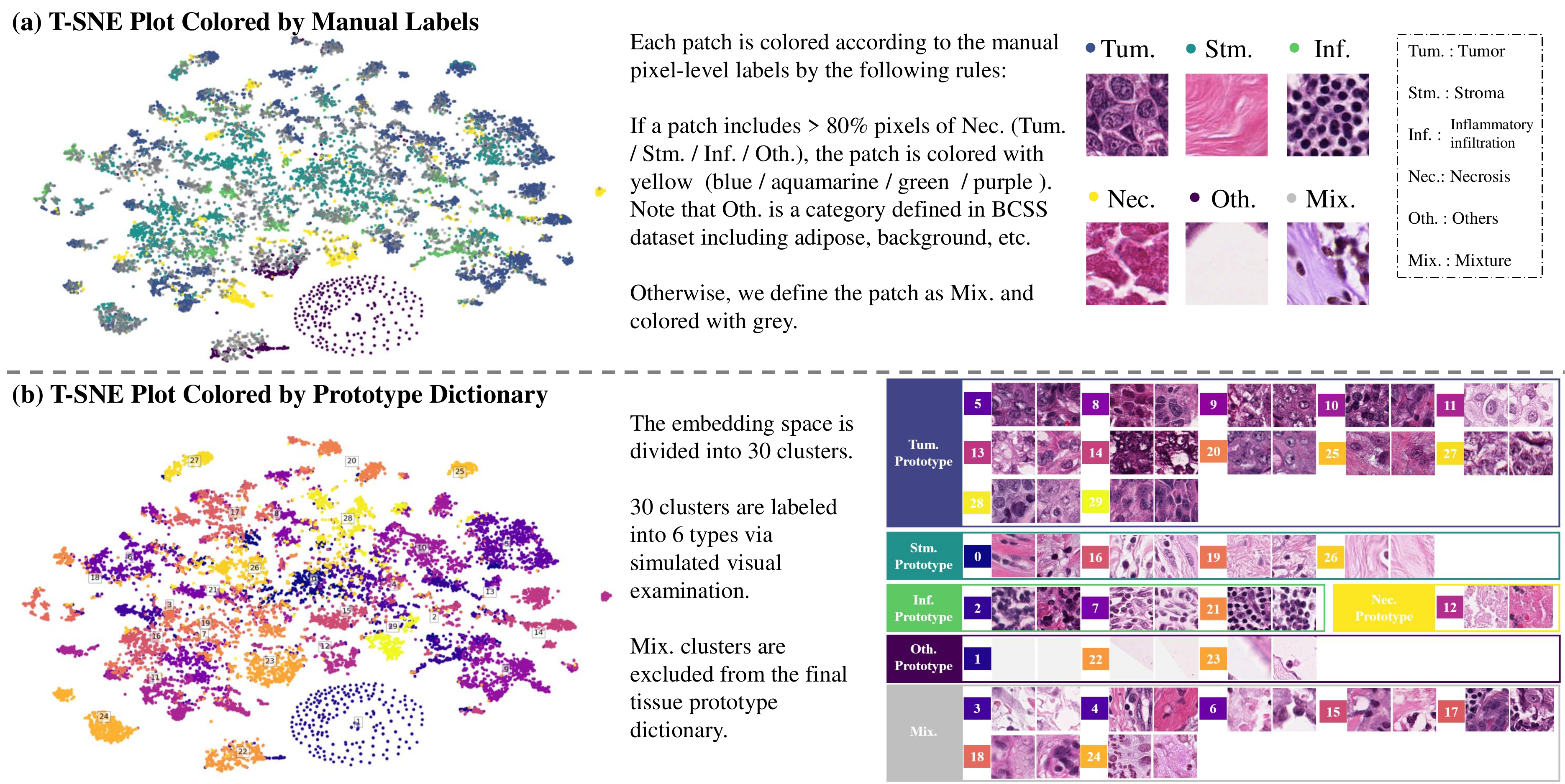}
\caption{Visualization of BCSS patches sampled for dictionary building in the embedding space via T-SNE and colored with: (a) manual pixel-level and (b) our clustering-based prototype-level labels. We also visualize 2 representative patches for each cluster.}
\label{fig_emb}
\end{figure*}

To help readers better understand, we in Fig. \ref{fig_emb} illustrate the dictionary building process with the BCSS dataset \cite{amgad2019structured}. We visualize the randomly sampled 20000 patches used for dictionary building with T-SNE in Fig. \ref{fig_emb} (a) which is colored according to the manual pixel-level labels. Particularly, these patches are divided into 6 types, of which 5 types (tumor, stroma, inflammatory infiltration, necrosis, and others) are defined by the dataset providers, and the mixture type defined by us. Our tissue dictionary is built on the clustering results drawn in Fig. \ref{fig_emb} (b) by K-Means++ with $k=30$. Intuitively, we can observe that the clustering results maintain a high degree of consistency with the manual labels. By sampling 10 representative patches from each cluster center and calculating the tissue proportion, we can find that 13/5/3/1/3 clusters are identified as tumors/stroma/inflammatory infiltration/necrosis/others, correspondingly. Note that Cluster 1 is formed by blank background patches, therefore spreading in a centrosymmetric distribution. There are 5 cluster with multiple tissue mixture patterns, which are excluded from the dictionary.

After above process, we can build the prototype dictionary $\mathcal{C}$ with $l$ prototypes after the above process. The prototype dictionary can be noted as  $\mathcal{C}=\{c_1:y_1, c_2:y_2, ..., c_l:y_l\}$, where $\{c_1, c_2, ..., c_l\}$ are prototype embeddings, and $\{y_1, y_2, ...y_l\}$ are prototype-level labels given by the pathologist.

\subsection{Coarse Segmentation Prediction and Refinement}
Having obtained the pre-trained $f_\omega(\cdot)$ and prototype dictionary $\mathcal{C}$, we can generate coarse segmentation masks for a given WSI $i \in \mathcal{R}^{H \times W \times 3}$ by the process shown in Fig. \ref{fig_framework} Step 3.  
We first feed WSI $i$ into the encoder $f_\omega(\cdot)$ to have a $v \in \mathcal{R}^{\frac{H}{32} \times \frac{W}{32} \times 512}$ semantic vector map.
Then, we can query $\mathcal{C}$ to determine the tissue type for each location. Intuitively, for each location, we can directly retrieve the nearest prototype to the current semantic vector from $\mathcal{C}$ and label each location with the corresponding tissue type. We termed such a query setting as Direct-Query (DQ). However, DQ treats each location independently ignoring intrinsic similarity between the embeddings. Thus, we further improve DQ with an enhanced Cluster-then-Query (CQ) setting to suppress outlier query results. Particularly, in order to explore the intrinsic similarity between the embeddings in the given feature map, we divide the $\frac{H}{32} \times \frac{W}{32}$ locations into different groups by performing K-Means++ on the $\frac{H}{32} \times \frac{W}{32}$ 512-d vector set. After that, the clustering centroids of different groups are used to query $\mathcal{C}$ for labeling each location. As for the clustering number of the target WSI, we find that it is better to decide the clustering number according to tissue types in the given WSI. According to the DQ based query results, we can roughly know that there may exist $d$ tissue types for the target WSI. Considering the possibility of spatial dispersion of the same tissue in images, We divide the current feature map into $\gamma \times d$ clusters. $\gamma \geq 1 $ is the scale factor and set to 5 empirically. 

Now, we have a $\frac{H}{32} \times \frac{W}{32} \times 1$ tissue segmentation map, which can be upsampled to restore its resolution for the given WSI. The coarse map has zigzag edges but it can offer dense supervision for segmentation network training. To refine the final prediction, we further use these coarse masks as the pseudo tissue masks to train a segmentation network from scratch. 
By such, we finally build a bridge from the prototype learning to semantic segmentation.

\section{Experiments}
\label{sec:exp}

\begin{table*}[t]
\caption{Quantitative segmentation results of different strategies. Tum./Nor./Stm./Inf./Nec./Oth. is short for Tumor/Normal/Stroma/Inflammatory infiltration/Necrosis/Others. DQ/CQ stands for the Direct-Query/Cluster-then-Query setting, correspondingly The best results are addressed in bold. Note that ABMIL can only be used for the binary prediction task.}
\label{tab_1}
\centering
\scalebox{0.6}{
\begin{tabular}{c|c|c c c|c c c c c c}
\hline
\multicolumn{2}{c}{Dataset} & \multicolumn{3}{|c|}{{ CAMELYON16}}  & \multicolumn{6}{c}{{ BCSS}}  \\ \hline 
\multicolumn{2}{c|}{Metrics } & { Pix. Acc.} & { Tum. Dice} & { Nor. Dice} & { Pix. Acc.} & { Tum. Dice} & { Stm. Dice} & { Inf. Dice} & { Nec. Dice} & { Oth. Dice} \\ \hline 

\multicolumn{2}{c|} {Pixel-level Supervised LinkNet} & {0.9391} & {0.9078} & {0.9896} & {0.8335} & {0.8773} & {0.8162} & {0.7662} & {0.7640}  & { 0.9653}  \\ \hline 
\multirow{6}{*}{Coarse Seg.} & {Grad-Cam} & {0.8243} & {0.7480} & {0.9745 } & {0.5799} & {0.4415} & {0.6190} & {0.5122} & {0.3251}  & {0.7244}  \\ 
{} & {Grad-Cam++} & {0.7462} & {0.6169} & {0.9641} & {0.6121} & {0.6327} & {0.6041} & {0.3606} & {0.2757} & {0.7638} \\ 
{} & {ABMIL} & {0.8455} & {0.8033} & {0.9805}  & { N/A }  & { N/A } & { N/A} & { N/A} & { N/A} & {N/A}    \\ 
{} & {CLAM} & {0.8700} & {0.7752} & {0.9743} & {0.6441} & {0.7082} & {0.6256} & {0.6404} & { 0.3571} & {0.9109}    \\ 
{} & {Our Proto2Seg(DQ)} & {0.9174} & {0.8401} & {0.9810} & {0.7421} & {0.7907} & {0.7240} & {0.6590} & {0.6176}  & {0.9430}  \\ 
{} & {Our Proto2Seg(CQ)} & {0.9176} & {0.8491} & {0.9824} & {0.7484} & {0.7956} & {0.7281} & {0.6576} & {0.6381} & {0.9429}  \\ \hline

\multirow{3}{*}{Refine. Seg.} & {CLAM + LinkNet} & {0.9189} & {0.8741} & {0.9859} & {0.7323} & {0.7797} & {0.7281} & {0.6784} & {0.6337} & {0.9054}\\ 
{} & {Our Proto2Seg(DQ) + LinkNet} &\textbf{0.9266} & {0.8766} & {0.9859}  & {0.7491} & {0.8029} & {0.7604} & \textbf{0.6800} & {0.6143}  & \textbf{0.9479}\\ 
{} & {Our Proto2Seg(CQ) + LinkNet} &{0.9231} & \textbf{0.8824} & \textbf{0.9868}  & \textbf{0.7750} & \textbf{0.8063} & \textbf{0.7769}  & {0.6799} &  \textbf{ 0.6979} & {0.9477}\\ 
\hline
\end{tabular}}
\end{table*}

\subsection{Setup}
\textbf{Datasets:} Our experiments are conducted on two public datasets: \textbf{CAMELYON16:} we use the train set which contains 270 WSIs focusing on sentinel lymph nodes with pixel-wise cancerous/normal region annotation \cite{bejnordi2017diagnostic}. \textbf{BCSS}: contains 151 Region-of-Interest cropped Images from breast cancer WSIs in TCGA with tissue-level annotations \cite{amgad2019structured}. Five tissue types are labeled: tumor, stroma, inflammatory infiltration, necrosis, and `others' (including background, blood, etc). 
Similar to previous works \cite{xu2019camel,yan2022deep}, we generate the 2048$\times$2048 / 1024$\times$1024 image-level data from CAMELYON16/BCSS with the specimen-level pixel size of $1335nm$/$250nm$ to get a proper view of biological tissues as the image-level data. For BCSS, we follow the official train-test split with 2151/976 images for training/testing. For CAMELYON16, we randomly divide patients into the train-test set by 2:1 following the hold-out setting and have 1348/848 images for training/testing.

\noindent\textbf{Baselines \& Evaluation:} 
We compare our framework with the pixel-level supervised upper bound established by training a LinkNet \cite{chaurasia2017linknet} with ground-truth annotations, 
and several weakly supervised baselines including Grad-Cam \cite{selvaraju2017grad}, Grad-Cam++ \cite{chattopadhay2018grad}, ABMIL \cite{ilse2018attention}, and CLAM \cite{lu2021data}.  
The $2048\times2048/1024\times1024$ images from Camelyon16/BCSS are cropped into $128\times128$ patches for MIL methods. All networks are built based on ResNet18. We report macro-average pixel accuracy and each tissue's Dice score to compare different methods.

\noindent\textbf{Our Protot2Seg:} 
We first crop the training set into the $128\times128$ patches without overlapping resulting in over $330k/140k$ patches for CAMELYON16/BCSS, respectively. All the patches are used to train a ResNet18-based encoder following the original SimCLR \cite{chen2020simple} setting for 200 epochs. For computation efficiency in the clustering process, 20000 patches are randomly sampled for tissue prototype clustering on both datasets. Note that, the testing set is not exposed to the encoder during the training or the dictionary building process. We set the tissue cluster number of CAMELYON16/BCSS as 15/30 through the elbow method. We employ prototype dictionary and CQ setting to generate coarse segmentation. The training of the refinement segmentation network follows the same setting as our pixel-level supervised baseline except for using coarse segmentation masks as the supervision. 

\subsection{Main Results}

\begin{table*}[t]
\caption{Labeling Cost of different annotation types with BCSS training dataset. }
\label{tab_2}
\centering
\scalebox{0.8}{
\begin{tabular}{c|c|c|c|c }
\hline
{Annotation Type} & {Quantity}  & {Average Time }& {Total Time } & Ratio  \\ \hline 
{Pixel-level } & {2151 images} & { 8min50s per image }& 316.7h & 1583.5x\\ 
{Image-level } & {2151 images} & { 1min59s per image }& 71.1h & 355.5x  \\ 
{Prototype-level (ours) } & {30 clusters} & { 26s per cluster }& 0.2h & 1x \\ 
\hline
\end{tabular}}
\end{table*}

\noindent\textbf{Quantitative results:} 
The results of segmentation performance are summarized in Table. \ref{tab_1}. Our Proto2Seg(DQ) coarse segmentation results obtained via directly querying the tissue prototype dictionary already surpass other weakly supervised baselines with significant margins. By using the Proto2Seg(CQ) coarse masks as dense supervision, the refined segmentation results achieve comparable pixel accuracy and Dice as the pixel-level supervised upper bound on CAMELYON16 for binary segmentation. For BCSS, most tissue segmentation results are further improved by Proto2Seg(DQ) + refinement training, except for the necrosis. We conjecture that the coarse masks of the necrosis provide inaccurate supervision with only 1 prototype as shown in Fig. \ref{fig_emb} (b) and mislead the refinement. Our Proto2Seg(CQ) setting, taking the embeddings' intrinsic similarity and the tissue's spatial dispersion into consideration, further improves the coarse segmentation performance. We can also observe that the CQ based coarse segmentation masks offer better supervised signals than masks generated by DQ. We also enhance the CLAM segmentation results with the refinement training step, and we can find that our solution is still superior to the CLAM + refinement. In conclusion, our solution can narrows the performance gap between pixel-level fully-supervised and weak-supervised methods with prototypes.  

Note that we achieve better segmentation results with a lower annotation burden than other methods. 
To compare the labeling cost, we invited 3 pathologists to annotate randomly sampled 20 BCSS patches with 1024x1024 pixels for pixel/image-level labels and use the average time to measure the labeling cost for BCSS training set reported in Table. \ref{tab_2}. Obviously, our method is much more labor-saving. It is also reminded that we used 1024 $\times$ 1024 cropped BCSS images but the original WSIs can be with a resolution of $>$10000 $\times$ 10000, i.e. it will be more time-consuming to give pixel/image-level labels for the original WSIs. But our method will not be affected by larger resolutions.

\begin{figure*}[t]
\centering
\includegraphics[width=1\columnwidth]{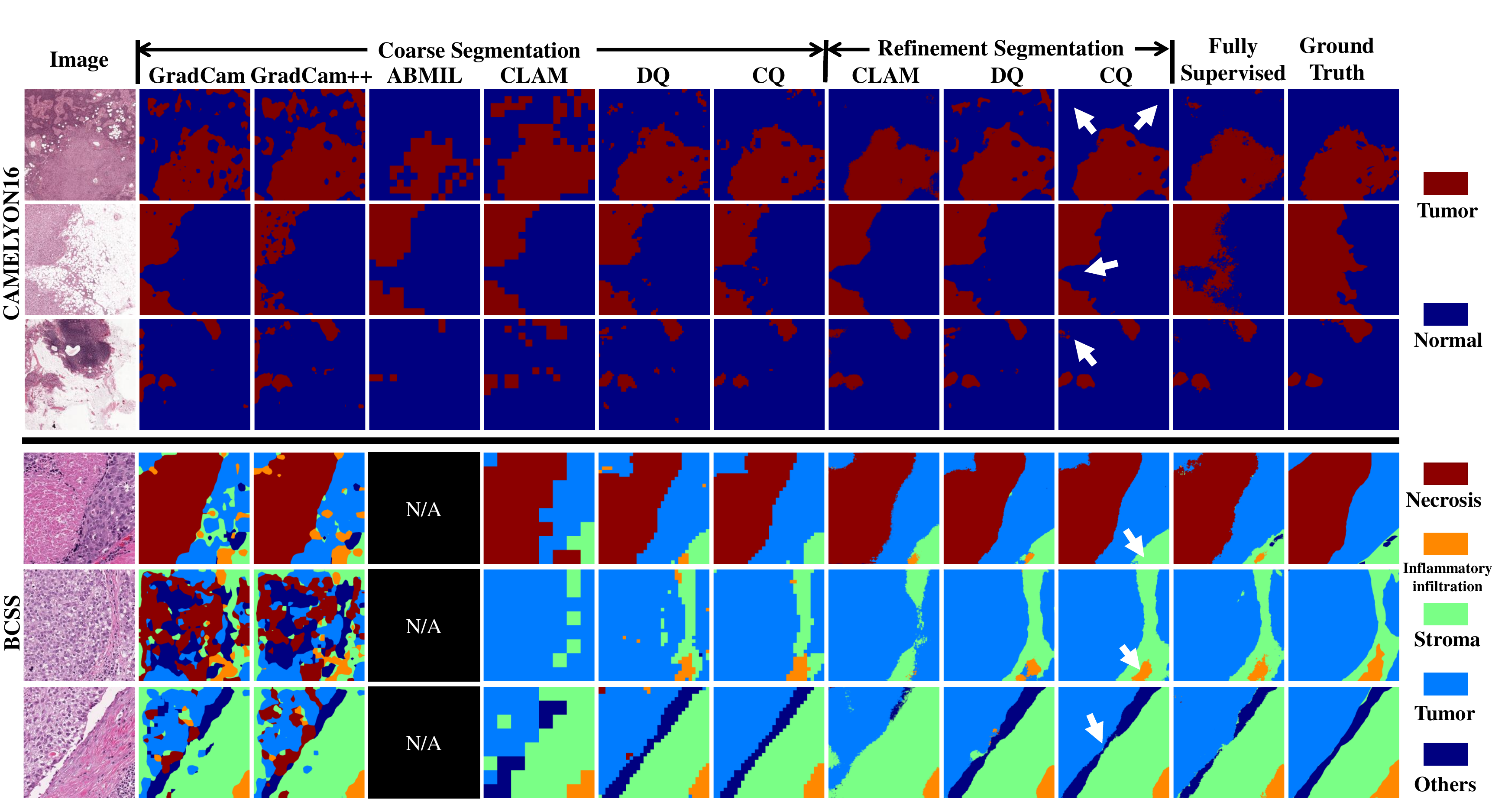}
\caption{Qualitative comparison of different methods. DQ/CQ stands for the Direct-Query/Cluster-then-Query setting, correspondingly. White arrows address where the coarse segmentation results are improved by the CQ setting on CAMELYON16/BCSS.}
\label{fig_comparison}
\end{figure*}

\noindent\textbf{Qualitative results:}  We visualize the qualitative results in Fig. \ref{fig_comparison}. We can find that our methods can better handle binary and multi-tissue segmentation than other weakly-supervised methods. Because the DQ setting treats every location independently, there exist outlier errors in its coarse segmentation results. We use white arrows to address where the coarse segmentation results are further improved by our CQ strategy. By using the coarse segmentation as the dense supervision, it can be observed that the fine results better correct some error predictions and have smoother edges. 

\subsection{Ablation Study}

\begin{figure}[h]
\centering
\includegraphics[width=1\columnwidth]{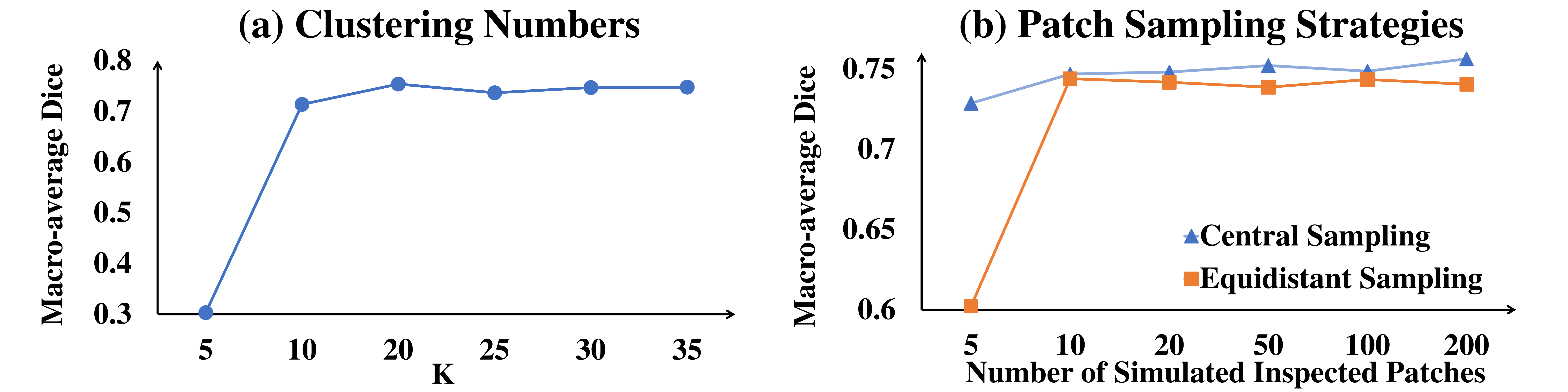}
\caption{Ablation studies on (a) clustering numbers and (b) representative patch sampling strategies on BCSS. The macro-average Dice of 5 tissues is reported.}
\label{fig_ablation}
\end{figure}

\noindent\textbf{Clustering Numbers for Prototype Identification:} 
Fig. \ref{fig_ablation} (a) illustrates how the selection of cluster number $k$ for dictionary building can affect the results. 
When the number of clusters $k<10$, it's difficult for our method to distinguish different tissues, because too few clusters may not capture enough inter-/intra-class heterogeneity. When $k>10$, the coarse segmentation masks can obtain satisfied segmentation results. As mentioned above, we follow the elbow method to set $k=30$ without accessing any labels, while the optimal Dice is achieve when $k=20$. That is, determining proper $k$ is an open challenge for unsupervised clustering, but our method is robust when $k>10$.

\begin{table*}[t]
\caption{The effects of different clustering algorithms on CAMELYON16 dataset. }
\label{tab_cluster}
\centering
\scalebox{0.8}
{
    \begin{tabular}{c|c|c|c|c|c}
         \hline
         Clustering Method & Kmeans & Kmeans++ & SpectralClustering & DBSCAN & FINCH\\ \hline
         Time (s) & 12.82 & 10.67 & 108.9 & 10.85 & 8.08 \\ \hline
         Mean Dice (DQ) & 0.9000 & 0.9107 & 0.9045  & 0.7802 & 0.8942 \\ \hline
    \end{tabular}
}
\end{table*}

\noindent\textbf{Patch Sampling Strategies for Visual Inspection:} Fig. \ref{fig_ablation} (b) shows how the patch sampling strategy will affect the results on BCSS. We perform experiments about central/equidistant sampling strategies with different values of $t$ and report the macro-average Dice of 5 tissues. In equidistant sampling setting, we randomly sample $t$ equidistant patches in each cluster form center to border for inspection. We can conclude that central sampling is constantly better. We can also observe that checking too few patches leads to performance decrease while checking more patches leads to higher accuracy but with a  heavier workload. To make the trade-off, we employ central sampling with $t=10$.

\noindent\textbf{Different clustering algorithms:} Five clustering algorithms are evaluated with our framework shown in Table \ref{tab_cluster}. The clustering number of the former three algorithms are determined with the same elbow method described. DBSCAN\cite{ester1996density} and FINCH\cite{sarfraz2019efficient} are $k$-free, i.e. the algorithm can automatically determine clustering number. We can observe that Kmeans++\cite{arthur2006k} outperforms other methods. 

\section{Conclusion}
In this paper, we make one of the early attempts to bridge the contrastive learning-based WSI patch pre-training and semantic segmentation with a human-machine interactive labeling tissue prototype dictionary. 
Experiments on two public datasets demonstrate that our method is comparable to the supervised upper bounds and outperform other weakly-supervised methods.
The major limitations are that we simulate the visual examination of pathologists in current experiments and the coarse masks inevitably contain noises. Future work will discuss about how human factors may affect the visual injection process and incorporate techniques \cite{xu2022denoising,xu2021noisy,xu2022anti} to alleviate the negative impacts of label noises.

\subsubsection{Acknowledgement.}
This research was partly supported by the National Key R$\&$D Program of China (Grant No. 2020AAA0108303),  Shenzhen Science and Technology Project (Grant No. JCYJ20200109143041798), Shenzhen Stable Supporting Program (Grant No.  WDZC20200820200655001), and Shenzhen Key Lab of next generation interactive media innovative technology (Grant No. ZDSY
S20210623092001004).

\bibliographystyle{splncs04}
\bibliography{citation}

\end{document}